\documentclass[runningheads]{llncs}
\usepackage[T1]{fontenc}
\usepackage{graphicx}
\usepackage[english]{babel}
\usepackage{float}

\usepackage{blindtext}
\usepackage{subfiles}

\usepackage[export]{adjustbox}
\usepackage{wrapfig}
\usepackage{booktabs}
\usepackage{graphicx,subcaption}

\usepackage{amsmath}
\usepackage{mathtools}
\usepackage[thinc]{esdiff}
\usepackage{algpseudocode}
\usepackage{algorithm}
\usepackage{bm}

\usepackage{multirow}
\usepackage{tabularx}
\usepackage{makecell}
\makeatletter
\newcommand\notsotiny{\@setfontsize\notsotiny{6.31415}{7.1828}}
\makeatletter

\usepackage{scrextend}
\usepackage[printonlyused, nolist]{acronym} 
\usepackage{etoolbox}

\RequirePackage{doi}
\newcommand{\orcid}[1]{\href{https://orcid.org/#1}{\includegraphics[width=10pt]{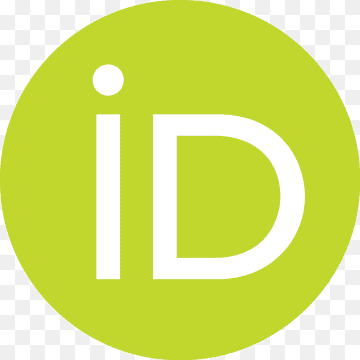}}}

\usepackage{hyperref}
\usepackage{cite}

\usepackage[colorinlistoftodos]{todonotes}

\usepackage{tikz}
\usetikzlibrary{shapes}
\usetikzlibrary{positioning}
\usetikzlibrary{math}
\definecolor{boxcolor}{rgb}{0.95, 0.95, 0.95}

\newcommand{\myedge}[4]{
    \begin{tikzpicture}[baseline=-2.5pt]
        \node[draw=gray,rounded rectangle,fill=boxcolor] (a)  at (0, 0) {\footnotesize{#1}};
        \node[] (b)  at (#4, 0) {\footnotesize{#2}};
        \node[draw=gray,rounded rectangle,fill=boxcolor] (c)  at (#4*2, 0){\footnotesize{#3}};
        \draw[gray] (a)--(b);
        \draw[gray] (b)--(c);
    \end{tikzpicture}
}

\colorlet{codecolor}{black!30}
\newcommand{\codebox}[1]{%
  \colorbox{codecolor}{\ttfamily \detokenize{#1}}%
}

\usepackage[shortlabels]{enumitem}

\begin{document}

%
\title{Graph Extraction for Assisting Crash Simulation Data Analysis}
%
%
\author{Anahita Pakiman\inst{1,2}\orcid{0000-0001-9706-7305} \and
Jochen Garcke\inst{1,3}\orcid{0000-0002-8334-3695} \and
Axel Schumacher\inst{2}\orcid{0000-0003-0649-5799}}
\author{Anahita Pakiman\inst{1,2}\orcid{0000-0001-9706-7305} \and
Jochen Garcke\inst{1,3}\orcid{0000-0002-8334-3695} \and
Axel Schumacher\inst{2}\orcid{0000-0003-0649-5799}}
\authorrunning{A. Pakiman, J. Garcke, and A. Schumacher}
%
\institute{Fraunhofer SCAI, Germany \\
\email{anahita.pakiman@scai.fraunhofer.de} \and
Bergische Universit\"at Wuppertal, Germany \and
Institut f\"ur Numerische Simulation, Universit\"at Bonn, Germany 
}
\maketitle              
\begin{abstract}
  In this work, we establish a method for abstracting information from Computer Aided Engineering (CAE) into graphs.
  Such graph representations of CAE data can improve design guidelines and support recommendation systems by enabling the comparison of simulations, highlighting unexplored experimental designs, and correlating different designs.
  We focus on the load-path in crashworthiness analysis, a complex sub-discipline in vehicle design.
  The load-path is the sequence of parts that absorb most of the energy caused by the impact.
  To detect the load-path, we generate a directed weighted graph from the CAE data.
  The vertices represent the vehicle's parts, and the edges are an abstraction of the connectivity of the parts.
  The edge direction follows the temporal occurrence of the collision, where the edge weights reflect aspects of the energy absorption.
  We introduce and assess three methods for graph extraction and an additional method for further updating each graph with the sequences of absorption.
  Based on longest-path calculations, we introduce an automated detection of the load-path, which we analyse for the different graph extraction methods and weights.
  Finally, we show how our method for the detection of load-paths helps in the classification and labelling of CAE simulations.


\end{abstract}

%
%
\section{Introduction}
\label{sec:intro}

We live in an interconnected world, and graph theory provides powerful tools for modelling and analysing this interconnectedness.
In graph theory, graphs are usually given in advance or easily abstracted from problems.
However, for many real-world scenarios, the individual data instantiations of modelled graphs need to be determined from the data before further analysis.
Therefore, the construction of high-quality graphs has become an increasingly desirable research problem, resulting in many graph construction methods in recent years \cite{qiao2018data}.
Furthermore, \ac{KG}s have become a new form of knowledge representation and are the cornerstone of several applications for specific use cases in industry.
The graph underlying the abstract structure, which effectively facilitates domain conceptualisation and data management, is the reason for the growing interest in this technology.
Moreover, the use of \ac{KG} is the direct driver of several artificial intelligence applications \cite{abu2021domain}.
Towards vehicle \ac{KG}, we aim to capture knowledge about vehicle development designs by automatically extracting graphs from a \ac{FE} model representing a vehicle.

The simplest scenario for identifying the connectivity of a graph is when it is associated with a physical problem related to the graph.
Such graphs include electrical circuits, power grids, linear heat transfer, social and computer networks, and spring-mass systems \cite{stankovic2020graph}.
In this work, we are interested in crashworthiness studies in vehicle design, where the transformation of crash simulation data into a graph is a challenging and unexplored area of research.
With the resulting representation, we aim to provide an abstraction of the problem that allows the use of graph theory methods for further automated analysis of the simulations.

\ac{CAE} analysis, mostly with the \ac{FEM}, enables car manufacturers to analyse many design scenarios, nowadays between 10,000 to 30,000 simulations per week \cite{Schwanitz2022}.
In crashworthiness analysis, \ac{CAE} engineers optimise the distribution of impact energy in the vehicle structure to reduce injuries to occupants or vulnerable road users.
How to characterise the sequence of absorbed energy, known as the load-path, is a fundamental question in this analysis.
The results of crash simulations include several outputs, such as deformations, accelerations and internal energy.
However, the load-path is not explicitly calculated in a crash simulation.
Therefore, a \ac{CAE} engineer must visualise the sequence to reveal the load-path. In this work, we propose and investigate graph representations for an automated identification of the load-path from the simulation data.

We consider parts of the \ac{FE} model entities as vertices of the structural graph following the scheme of \cite{pakiman2022graph}.
We want to detect the graph edges that resemble the structural connectivity of the vehicle.
We propose three approaches to determine this structural graph: \ac{CBG}, \ac{sPBG} and \ac{mPBG}.
The \ac{CBG} follows two steps: finding the connection of the components (a group of parts) and then identifying the connection of the parts in each component.
The \ac{sPBG} and \ac{mPBG} graphs have additional steps to convert the component connections to part connections, which requires the detection of the parts that are entangled in the connection that is supporting the flow of energy.

Defining the vehicle structure as a graph is the first step in load-path detection.
Secondly, we compute it as the longest path in weighted directed graphs, where the edge weights between the parts shall represent the energy flow during the crash.
We study different edge weighting functions for three graph extraction scenarios
and analyse the determined load-paths from an engineering perspective.
In this work, the investigation is carried out on the frontal structure of a complete vehicle with a multi-scenario load-path in a full frontal load case.
But, our approach is applicable to different impact directions and load case scenarios.

In summary, the main contributions of this work are:
\begin{itemize}
    \item the conversion of a vehicle structure to a weighted directed graph,
    \item the extraction of features representing the energy flow,
    \item a further graph segmentation that captures the time sequence of events,
    \item an automated detection of the load-path,
    \item the clustering of simulations based on their load-paths.
\end{itemize}

%
%
\section{Related work}
\label{sec:rw}

Recently, a graph schema to model vehicle development with a focus on crash safety was introduced in~\cite{pakiman2022graph}.
The graph modelling considers the CAE data in the context of the R\&D development process and vehicle safety, with the aim to enable searchability, filtering, recommendation, and prediction for crash CAE data during the development process.
In~\cite{pakiman2022graph}, the car parts are directly connected to their simulation, and the parts between the simulations have a connection to similar design based on the properties ID (PID) of the parts.
But, connections between the parts of one simulation are missing, therefore the vehicle's structure and its connectivity is not modelled.
Thus, incorporating the vehicle structure into the graph structure will enrich the data representation.

In crashworthiness, graphs have been used to predict the response of the vehicle \cite{granda2011automating} or barrier \cite{granda2016bond} with so-called bond graphs.
The bond graphs available for vehicle crashes represent the problem from the perspective of a mass-spring model \cite{granda2011automating}.
Bond graphs are ideal for visualising the essential properties of a system because their graphical nature separates the system structure from the equations. \cite{gawthrop2007bond}.
Bond graphs represent the vehicle structure by summarising the physical elements and connections.
However, to the best of our knowledge, there is no way of automatically extracting the vehicle structure as a bond graph.

Before the growth of computing power allowed large \ac{FEM} analysis, there were other modelling techniques that simplified the problem to a mass-spring model.
The advantage of the mass-spring model is that it can be easily represented as a weighted graph.
SISAME (Structural Impact Simulation And Model-Extraction) is a general-purpose tool for the extraction and simulation of one-dimensional non-linear lumped parameter structural models \cite{mentzer1992sisame}.
Using SISAME, mass element weights and spring element load-paths were optimally extracted directly from the test data accelerations and wall forces \cite{lim2017lumped}.
However, the \ac{LMS} modelling is one-dimensional and focuses mainly on accurately modelling the test data rather than representing the structural performance of the vehicle.
Later the \ac{DSM} model was introduced \cite{lange2019early} to compensate for the limitations of the \ac{LMS}.
It can only roughly capture displacements and energy absorption, neglecting connections and interactions with other components.

Another use of graphs in crash analysis is in the structural optimisation of the vehicle \cite{ortmann2013graph, schneider2017finding}.
Here, the optimisation method adds vertices and edges to stiffen the structure, starting with a simple graph describing the perimeter of the vehicle.
The focus of these studies is to search with a graph for the optimal solution of the vehicle design.
As a result, to complete the vehicle design and ensure safety performance, further processes and \ac{CAE} analysis are required.

To summarise, automatically converting a crash \ac{FE} model in vehicle development to a graph is still an open research question.
Depending on the detail required in a graph, there are several ways to represent an \ac{FE} model of a vehicle.
As a specific application, we investigate how adding connections to the graph will allow a load-path analysis for each simulation.
For that, we use and extend the recently introduced energy absorption features~\cite{pakiman2022knowledge}, which characterize the simulation's behaviour, as edge features to enable the load-path detection.

%
%
\section{Graph extraction}
\label{sec:graph_gnrtn}

It is a challenging task to generate a graph representing the structure of a vehicle from \ac{CAE} data.
Finding the connectivity of the parts is complex due to the number of connections, the variety of \ac{FE} modelling techniques and the variety of physical types of connections.
The best way to obtain this information would be to use the \ac{CAD} database, which is more standardised than \ac{CAE}.
However, this data depends on the company's workflow to maintain the link between the \ac{CAE} and \ac{CAD} models, which has yet to be well established.
In addition, these databases lack information on the dependencies of the part connections, i.e. all parts are connected without any hierarchy.
This hierarchy is essential for defining the direction of the edges and for identifying the vertices of the graph as either dead ends or capable of allowing energy to flow through the structure.
As a result, we are looking for a method to perform this intelligently using the \ac{FE} model, based on the location and closeness of parts therein.

The \ac{FE} model contains mesh faces and volumes with different entities representing the connections.
The mesh is defined by nodes and elements, where the element size defines the resolution of the discretization.
The nodes can represent the vertices and the elements define the edges for a graph defined as $G(V,E)$ with vertices and edges.
Consequently, a \ac{FE} model mesh itself represents a graph.
However, this graph has drawbacks.
A small element size, three to five $mm$, for a complete vehicle will result in a large number of vertices, up to 20 million, which is computationally expensive for \ac{GML} and the lack of semantics makes it difficult to analyse engineering concepts.
Coarsening the crash \ac{FE} mesh is an alternative, which is a topic in \ac{FE} modelling~\cite{bank1996algorithm, chawla2004mesh, montevecchi2017finite}.
However, rather than focusing on post-processing aspects, these studies have mainly focused on reducing the compute time of the \ac{FE} simulation.
Nevertheless, the result will still be a disconnected graph because a \ac{FE} model contains multiple meshes whose connectivity is not element based.
Therefore, we focus on linking \ac{FE} entities to extract the structure of a vehicle as a connected graph.

To determine the connectivity, we split the graph extraction problem into two steps.
First, component-level connectivity and then connectivity of parts within a component.
Thereby we keep hierarchy information in the graph structure.
Previously, we introduced a grouping method for identifying components \cite{pakiman2023smrnk}.
Here, we extend this method to search for connections between components.
In addition, we add edges to the graph that connect parts that belong to the same component.
To include timing in the graph, we also investigate to add a timing segmentation based on the timing of outgoing edges, see Section \ref{subsec:smgnt_time}.

\begin{figure}[t]
  \centering
  \begin{subfigure}[b]{0.35\linewidth}
    \includegraphics[width=\linewidth]{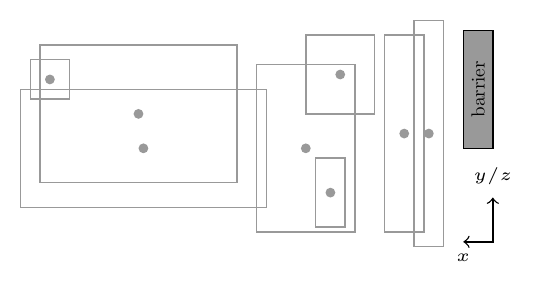}
    \caption{Each part as a box with its centre of gravity}
    \label{fig:mg0}
  \end{subfigure}%
  \hfill
  \begin{subfigure}[b]{0.3\linewidth}
      \raisebox{0.14\height}{\includegraphics[width=\linewidth]{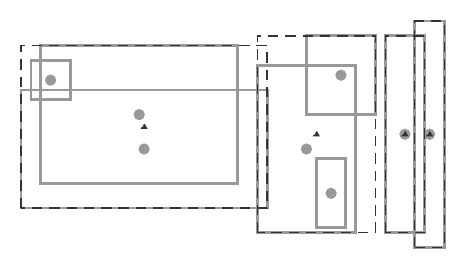}}
    \caption{Group parts as a component}
    \label{fig:mg1}
  \end{subfigure}%
  \hfill
  \begin{subfigure}[b]{0.3\linewidth}
    \raisebox{0.14\height}
    {\includegraphics[width=\linewidth]{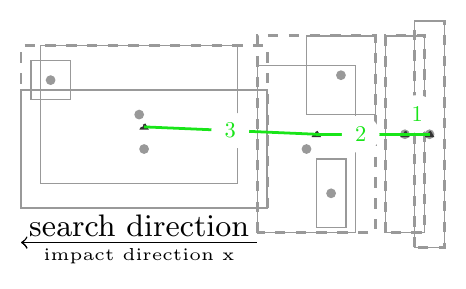}}
    \caption{Edges connecting components}
    \label{fig:mg2}
  \end{subfigure}%
  \hfill
  \begin{subfigure}[b]{0.3\linewidth}
    \includegraphics[width=\linewidth]{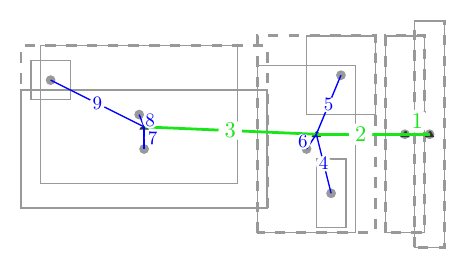}
    \caption{\ac{CBG}}
    \label{fig:mg3}
  \end{subfigure}%
    \hfill
  \begin{subfigure}[b]{0.3\linewidth}
      \includegraphics[width=\linewidth]{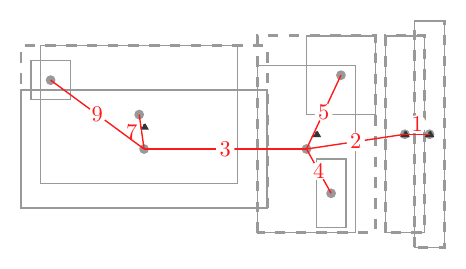}
      \caption{\ac{sPBG}}
      \label{fig:mg41}
  \end{subfigure}%
  \hfill
\begin{subfigure}[b]{0.3\linewidth}
    \includegraphics[width=\linewidth]{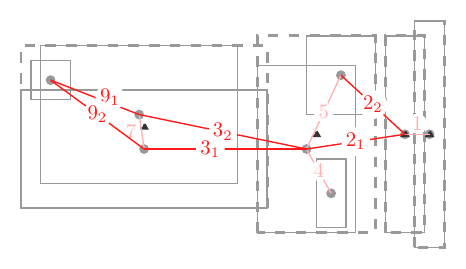}
    \caption{\ac{mPBG}}
    \label{fig:mg42}
\end{subfigure}%
    \caption{Abstracted visualization of the stages for graph extraction. While the method works in 3D, we here show a 2D visualisation. Solid squares: part, dashed square: component, circle: part box \ac{COG}, triangle: component box \ac{COG}, green edges: component to component, blue edges: component to part, red edges: part to part.}
    \label{fig:nrg}
\end{figure}

We consider the parts of the \ac{FE} entities as vertices of the structural graph of the vehicle, which follows the scheme of \cite{pakiman2022graph}.
We want to detect the edges that resemble the structural connectivity of the vehicle, and we propose three scenarios to do this: \ac{CBG}, \ac{sPBG} and \ac{mPBG}.
We need to extract information from the structure of the vehicle to obtain the connectivity between parts.
To do this, we create 3D axis-aligned boxes for each part that contain the volume of the part's geometry, Figure \ref{fig:mg0}.
Then, based on the overlap of the boxes, we define rules to group them as components, Figure \ref{fig:mg1}, and later form the structure of the graph from the overlap of the boxes.
In the following three subsections, we will discuss the detailed differences between these methods and for now only describe the general idea.
The \ac{CBG} follows two steps: finding the connections between components, Figure \ref{fig:mg2}, and then determining the part connectivity in each component, Figure \ref{fig:mg3}.
\ac{sPBG} and \ac{mPBG} have additional steps to convert component connections to part connections, which requires identifying the parts involved in the connectivity that supports the energy flow.
We explore two scenarios for this as single and multi-part-based graphs, Figures \ref{fig:mg41} and \ref{fig:mg42}, respectively.
For all these methods, we consider a directed graph whose directions are set to have a positive inner product with the impact axis, direction $x$ in Figure \ref{fig:mg0} and Algorithm \ref{alg:cmpnt}.

\begin{algorithm}
  \caption{edge direction for vertices A and B, impact direction $x$}\label{alg:cmpnt}
  \textbf{Input:} TVL: Threshold Limit Value
  \begin{algorithmic}
    \If{$\|\bm{AB}\| < \bm{TLV}$}
      \If{$\overrightarrow{\bm{AB}} \cdot \overrightarrow{\bm{x}} > 0$}
        \State connect $\bm{A}$ to $\bm{B}$
      \Else
        \State connect $\bm{B}$ to $\bm{A}$
      \EndIf
    \EndIf
  \end{algorithmic}
\end{algorithm}

\subsection{\ac{CBG}}
\label{subsec:cbg}

The construction of \ac{CBG} requires first the detection of the components and then the detection of the connections between components.
The component detection considers each part to be a box, then groups them together as a component, and finally evaluates the component box.
For \ac{CBG}, in addition to the part vertices, we also introduce component vertices into the graph.
The location of these vertices is at the centre of the components and the component parts are connected to them.
For example, in Figure \ref{fig:mg0} with eight parts, four components are detected and corresponding component boxes are generated, in Figure \ref{fig:mg1}.
Then, using a threshold value ($TLV$), our algorithm searches for immediately adjacent components.
The thresholding allows having several neighbours.
The search algorithm sorts components by impact direction, starting from the impactor/barrier position and moving into the vehicle along the impact direction, e.g. $x$ in Figure \ref{fig:mg2}.
Finally, we connect all of the parts in each of the components to the component box.

The result at this stage, Figure \ref{fig:mg3}, is a connected graph, which is a heterogeneous graph of parts and components.
Evaluating the longest path for a heterogeneous graph requires additional evaluation of edge features between vertices of different types.
Therefore, our goal is to modify this graph into a homogeneous graph.
First, we consider only the components as vertices, delete the vertices of the parts, and evaluate the features of the component vertices based on the parts, as we introduced earlier in \cite{pakiman2023smrnk}.
This graph is \ac{CBG} and doesn't contain the detailed features of all the parts.
Another approach is to use the heterogeneous graph as an input to find further connectivities of the parts.
We explore this approach in Sections \ref{subsec:sPBG} and \ref{subsec:mPBG}.

\subsection{\ac{sPBG}}
\label{subsec:sPBG}

The \ac{sPBG} is a basic approach to convert the heterogeneous part-component graph into a part graph by transferring the component vertex and its corresponding edges to a part vertex.
Because of the single part selection, we call it \ac{sPBG} and we consider an alternative multiple part scenario in Section \ref{subsec:mPBG}.
There are several ways to determine the corresponding part for each component.
First, we use a simple scenario and select the largest part, the geometric aspect of the component, as the corresponding vertex for the component connection.
For example, in Figure \ref{fig:mg41} with this consideration, the \myedge{}{1}{}{.45} remains in the same position as the component-part graph because the connecting components contain a single part.
The edges \myedge{}{2,4,5}{}{.7} move from the component box to the largest part, so \myedge{}{6}{}{.45} is removed.
Finally, the edge \myedge{}{7}{}{.45} disappears in the last components and edges \myedge{}{8,9}{}{.55} move to the other end of edge seven.

The \ac{sPBG} graph is characterised by having a main connection from the beginning to the end of vehicles with several dead ends for each master part.
We expect that the identification of the energy flow of the simulation will be limited by the existence of many dead ends.
Furthermore, for \ac{sPBG} a single part is the representative of a component and therefore only a single part interacts with the other parts, which in some cases is not appropriate.
For example, the side-member, which is a thin-walled structure, has two U-sections welded and several reinforcement plates. 
In this example, information about the interactions of the other U-profiles and reinforcement plates will be missed if only one part is considered to represent the component.
Next, we consider multiple connections between the components with \ac{mPBG}.
Multiple connections reinforce the lack of internal connections compared to \ac{sPBG}.

\subsection{\ac{mPBG}}
\label{subsec:mPBG}

The \ac{mPBG} is an alternative to \ac{sPBG} by allowing multiple representatives for components.
This approach allows for part interactions in the components and between components.
Here we transfer and distribute the component vertices using the information from the component discovery process, rather than selecting the largest box.
As described in \cite{pakiman2023smrnk}, our component detection algorithm has two scenarios for identifying the components: full and partial overlap merge.
Full overlap means a box is completely within the parent box, whereas partial overlap addresses partially overlapping scenarios.
These two scenarios are treated differently for \ac{mPBG} extraction.
In the case of a full merge, the part is connected to its parent box, similar to \ac{sPBG}.
However, in partial overlap scenarios, both boxes will represent the component.
In this case, a component vertex is transferred to all partially overlapped boxes.
Nevertheless, each part will retain its connections to the child based on full merges.
Figure \ref{fig:mg42} visualises these two scenarios.
The edge \myedge{}{2,3}{}{.55} branches to two edges \myedge{}{$2_1$,$2_2$}{}{.7} and \myedge{}{$3_1$,$3_2$}{}{.7} respectively compared to the \ac{sPBG} due to a partial merge.
Furthermore, the edge \myedge{}{9}{}{.45} branches to \myedge{}{$9_1$,$9_2$}{}{.7} since it is added after the partial merge and belongs to both parent boxes.

%
%
\section{Load-path detection}
\label{sec:lp_dtc}

Understanding how an external load is transferred to a given structure helps to evaluate the performance of different components, improve structural strength and reduce structural weight in structural design and optimisation.
The so-called load-path of a component is a concept for tracking the transferred load within a structure, starting from the load points and ending at the support points, which has been studied in structural design for several years \cite{marhadi2009comparison}.
Reviews of different approaches to load-path detection are proposing a new metric to find detailed load-paths at mesh size for better component design.
However, we are interested in the load-path in the context of crash analysis, which involves the interaction of several components.
Load-paths are typically defined as vehicle parts capable of generating resisting forces during a crash event \cite{lindquist2003real}.
To identify load-paths during a crash, nine load-paths were first defined and classified in~\cite{lindquist2003real}.
These can be easily examined for signs of loading after a crash.
On the other hand, this work mainly introduces new measures for evaluating real crashes.

We aim to identify the load-path to be able to compare simulations by highlighting the importance of different paths during the crash.
We use the longest path calculation\footnote{The longest path in a directed acyclic graph, \codebox{dag_longest_path()}, from NetworkX} to find the load-paths involved in absorbing the crash energy.
In this calculation, we aim to look at the internal energy absorption of the parts since manufacturers optimise the energy absorption capabilities of the load-paths \cite{lindquist2003real}.
To achieve this, we use the so-called \ac{$IE$} features introduced in~\cite{pakiman2022knowledge}.
Initially, one has an unweighted graph with \ac{$IE$} features for vertices.
An essential step is to convert vertex features into edge weights.
In this way, the edge weights hold the absorption characteristics and instead of the longest unweighted path, we compute the potential load-path.

In the following subsections, we first introduce the edge weights as a single feature of the internal energy flow, $f_{IE}$, and the time segmentation, $s_t$.
$f_{IE}$ is computed from the vertices maximum absorbed internal energy ($IE_{max}$) using internal energy flow calculation, see Section \ref{subsec:flw_clctn}.
For $s_t$ we update the graph with time segmentation to have absorption time features on the edges, see Section \ref{subsec:smgnt_time}.
Finally, in Section \ref{subsec:cmbn_fts} we will present several ways to combine edge features.

\subsection{Internal energy flow}
\label{subsec:flw_clctn}
We consider the flow equation for the propagation of the \ac{$IE_{max}$} feature from the vertices to the edges, $f_{IE}$.
Our graph is a directed weighted graph $G(V, E)$ with vertices $V$, edges $E$ and a weight $w(e)$ assigned to each edge.
We assume that the energy flow from vertex $i$ to $j$, $w_{i,j}$, is represented by an edge weight between vertices $i$ and $j$.
The energy flow equation relates the absorbed internal energy $IE_j$ of a vertex $v_j$ to the balance of the input and output \ac{$IE$} from that vertex to its neighbours:
\begin{equation}\label{eq:flow0}
  IE_j =  \sum_{n \in I(j)}{w_{n, j}} - \sum_{n \in O(j)}{w_{j, n}}.
\end{equation}
For a vertex $v$ in a graph, we denote by $I(v)$ and $O(v)$ the set of in-neighbours and out-neighbours of $v$, respectively.
We start computing edge weights with vertices that only have incoming edges, called dead ends.
We compute the flow from the dead ends, backwards along their edge directions, to find the inflow of the dead ends vertices.
The active vertices for the next step calculation are the source vertices to the dead ends.
Consequently, if all their outflow energy is available, we can find the inflow energy to the active vertices.
Until all its outflows are known, a vertex is withheld from being an active vertex.
In addition, there is a different treatment for the dead ends at vertices that have an inflow degree of zero.
These source-only vertices reflect where the impact is initiated and where accordingly the kinetic energy input takes place.
Therefore, these vertices are not considered when they are marked as active vertices.
Instead, the edge weights of these source-only vertices are calculated when their outgoing neighbours are the active vertices.
In some cases, the weights of all their outgoing edges have already been evaluated, but the active vertices may have more than one incoming edge.
In this case, the energy flow is partitioned to the in-degree, $I(v)$.
An unequal stiffness of the structure does not allow an equal distribution.
Therefore, equal partitioning can lead to errors in the flow calculation, which we discuss in \ref{subsec:grph_flw}.

\subsection{Time segmentation}
\label{subsec:smgnt_time}

\begin{figure}[b]
    \centering
  \begin{subfigure}[b]{0.32\linewidth}
      \centering
        \includegraphics[width=\linewidth]{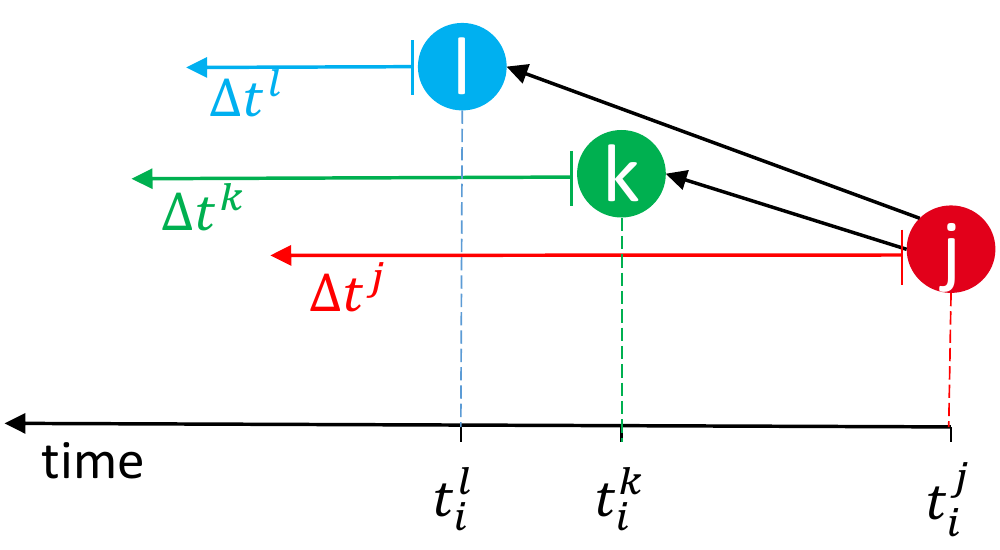}
        \caption{Initial state of the graph before time segmentation. }
        \label{fig:sgmnt_a01}
  \end{subfigure}%
\hfill
  \begin{subfigure}[b]{0.32\linewidth}
      \centering
        \includegraphics[width=\linewidth]{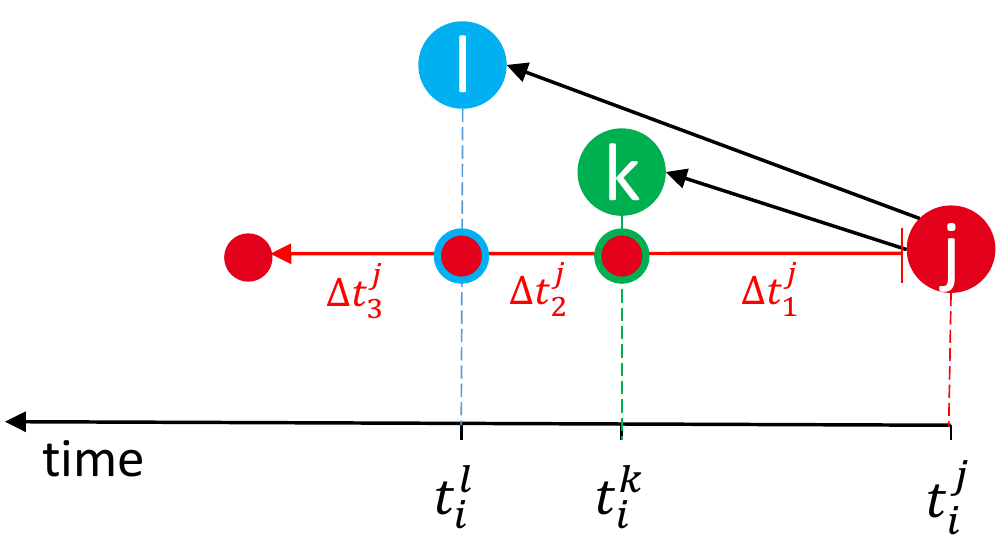}
        \caption{Adding vertices according to the successors. }
        \label{fig:sgmnt_a02}
  \end{subfigure}%
\hfill
  \begin{subfigure}[b]{0.32\linewidth}
      \centering
        \includegraphics[width=\linewidth]{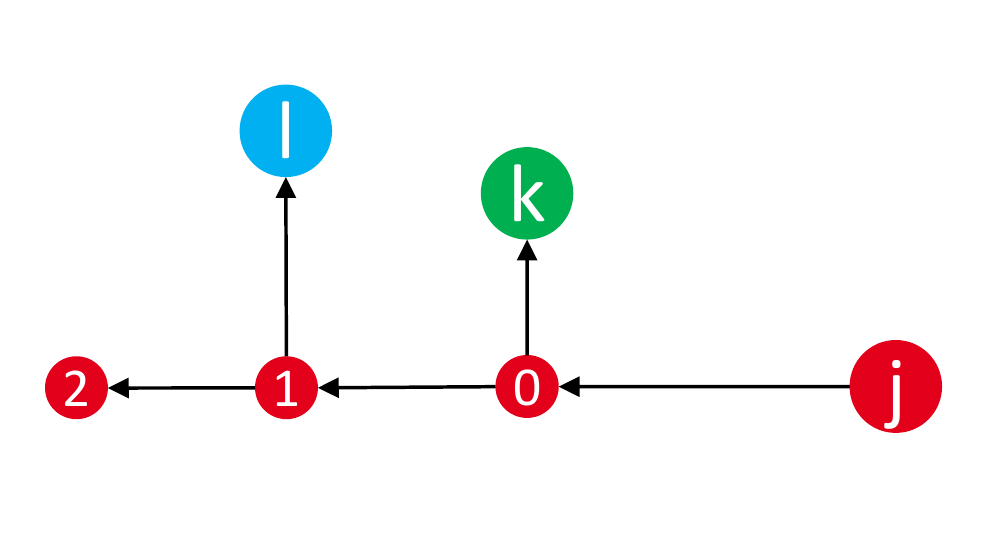}
        \caption{The graph after adding time segmentation vertices.}
        \label{fig:sgmnt_a03}
  \end{subfigure}%
    \caption{An example of time segmentation process for a vertex $j$ with two outgoing edges to the successors vertices of $k$ and $l$. The time axis shows the $t_i$ value for each vertex and absorption time with an arrow in front of each vertex. }
    \label{fig:sgmnt_abstrct}
  
\end{figure}

To convert the vertex absorption times into edge weights is more complex than the handling of \ac{$IE_{max}$}.
This is because the graph connectivity of the vertices differs from the time sequence of the parts that absorb energy.
Moreover, the time information of each vertex is an absorption interval ($\Delta t$), \ac{$t_i$} to \ac{$t_n$}, which may overlap with one of its neighbours.
In the example shown in Figure \ref{fig:sgmnt_abstrct}, we demonstrate the time segmentation for vertex $j$ with two successors of $l$ and $k$. 
In this figure, the absorption period of each vertex is plotted as a vector along the time axis. 
The overlap of these vectors highlights the need for time segmentation, see Figure \ref{fig:sgmnt_a01}.
To overcome this, we segment the time interval of the absorption for each vertex.
The segmentation is based on the $t_i$ value of the successors of the vertex.

Accordingly, we add vertices to the graph for each segmented time and connect each successor vertex to the vertex added for time segmentation.
In this example, a vertex is added to the graph for each successor vertex, $l$ and $k$, see Figure \ref{fig:sgmnt_a02}.
Note that if some of the successors have the same $t_i$, then only one vertex will be added.
In addition, to include the total absorption, an extra vertex is added to represent the total absorption vector as the sum of $\delta t^j = \delta t^j_1 + \delta t^j_2 + \delta t^j_3$, see Figure \ref{fig:sgmnt_a03}. 
Then we sort the $t_i$ of the successor vertices to find the connection between the new vertices.
Finally, the directed edges containing the time sequences and durations are added and the old edges are deleted, see Figure \ref{fig:sgmnt_a03}.
Additionally, we add the initial timing $t_i^k$ as a vertex feature for the kth segment, so that all vertices have a $t_i$.
Finally, the edge weight $s_t$ for time segmentation is for a directed edge from $m$ to $n$ defined by $s_t :=t^m_i - t^n_i$.

\subsection{Feature combination}
\label{subsec:cmbn_fts}

We consider two approaches to combine \ac{$IE_{max}$} and the timings of part absorption.
In the first approach, we modify the vertex feature \ac{$IE_{max}$} according to the absorption time before the flow computation from Section~\ref{subsec:flw_clctn}.
To do this we look at the integration of the $IE$ curve over time, $IE\Delta t$.
The start and end of the integration are set to the minimum $t_{i_{min}}$ and maximum $t_{n_{max}}$ of the absorption times, \ac{$t_i$} and \ac{$t_n$}, respectively, over all parts.
To simplify the calculation, we divide the area under the curve \ac{$IE$} into three zones.
For each zone the area under the curve, $A$, is calculated:

\begin{itemize}
  \item[$\circ$]$\ (t_{i_{min}},\ t_i)\ $\ \ unload period,\ \ \ \ \ \ \! $A_1=0$
  \item[$\circ$]$\ \ \ \ \ (t_i,\ t_n)\ $\ \ absorption period, $A_2 = IE_{max}(t_n -t_i)/2$
  \item[$\circ$]$(t_n,\ t_{n_{max}})\ $\ \ saturated period,\ \, $A_3 = IE_{max}(t_{n_{max}} -t_n)$
\end{itemize}

The sum of these areas is the new node feature and we compute, as in Section \ref{subsec:flw_clctn}, the combined edge weight with the flow of $IE\Delta t$, $f_{IE\Delta t}$.
In the second approach, we use the time segmentation graph.
For this graph,
we calculate the energy absorption efficiency, $P_e = IE/\Delta t$, where $\Delta t=s_t$, see Section \ref{subsec:smgnt_time}, and $IE=f_{IE}$, see Section \ref{subsec:flw_clctn}.

%
%
\section{Result}
\label{sec:reslt}

\begin{figure}[t]
  \centering
  \begin{subfigure}[b]{0.333\linewidth}
      \centering
        \includegraphics[width=\linewidth]{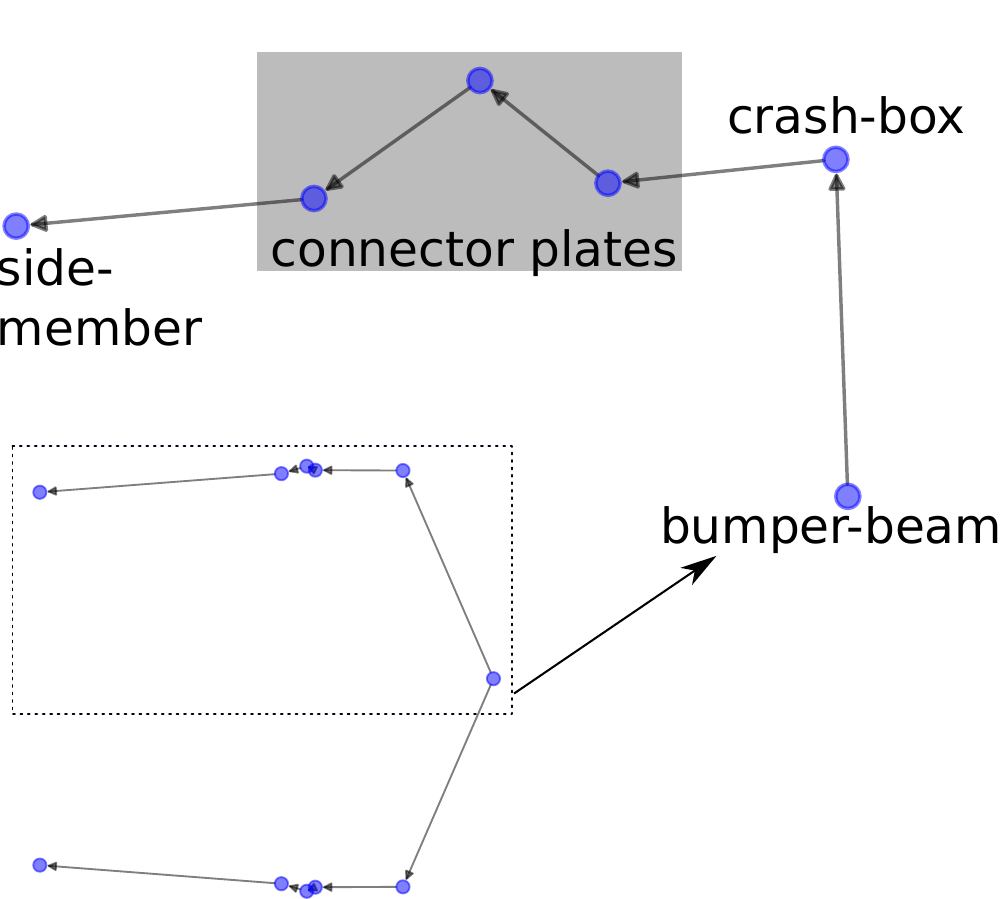}
        \caption{CBG}
        \label{fig:cmpnt}
\end{subfigure}%
\hfill
\begin{subfigure}[b]{0.333\linewidth}
    \centering
    \includegraphics[width=\linewidth]{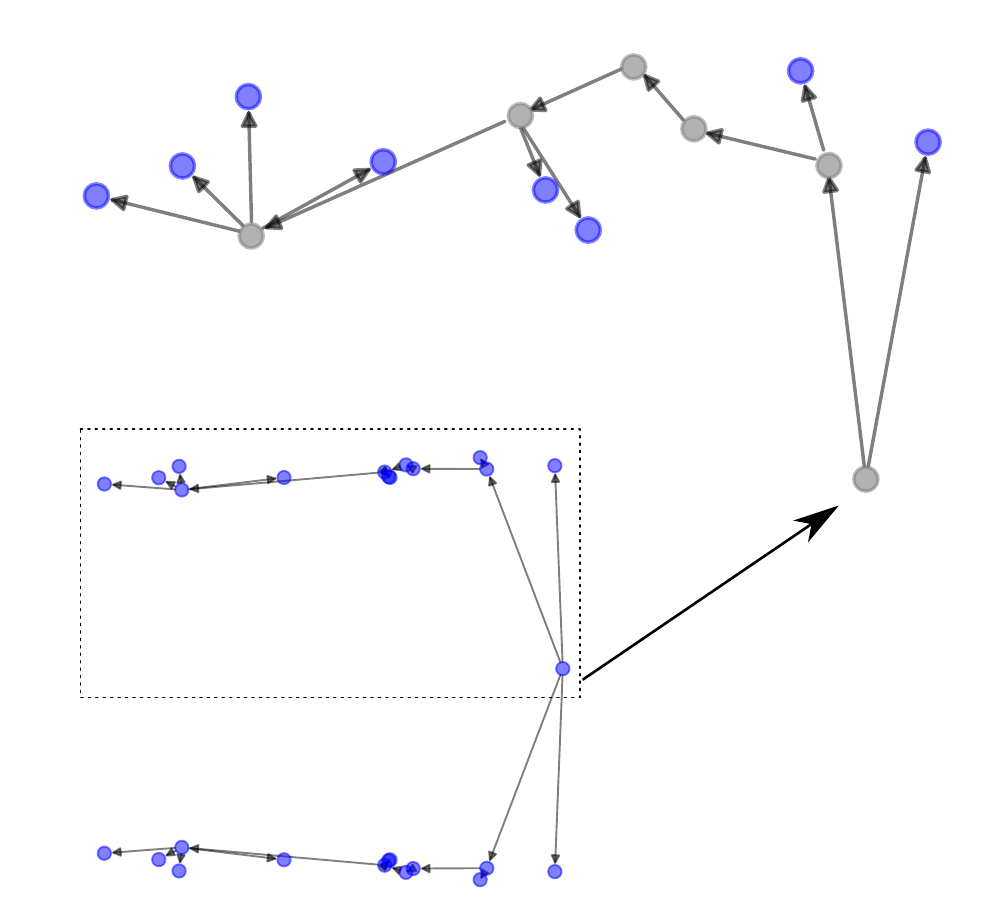}
      \caption{sPBG}
      \label{fig:open}
    \end{subfigure}%
    \hfill
      \begin{subfigure}[b]{0.333\linewidth}
          \centering
          \includegraphics[width=\linewidth]{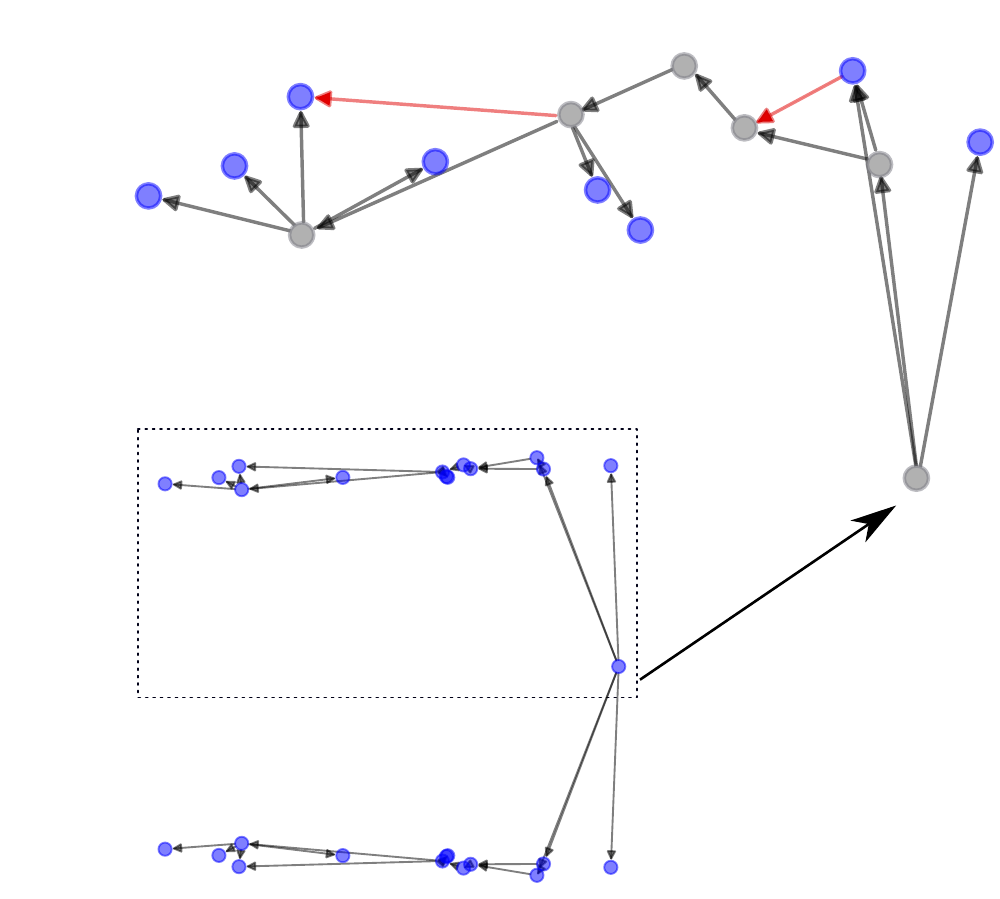}
            \caption{mPBG}
            \label{fig:looped}
\end{subfigure}
    \caption{Extracted graphs for the illustrative example\cite
    {pakiman2023smrnk}. A zoomed view of the upper half is shown for each graph\protect\footnotemark.
    The additional edges for \ac{mPBG} compared to \ac{sPBG} are marked as red in (\subref{fig:looped}).
    }
    \label{fig:extrct_g}
\end{figure}

\begin{figure}[t]
  \centering
  \begin{subfigure}[b]{0.5\linewidth}
      \centering
        \includegraphics[width=\linewidth]{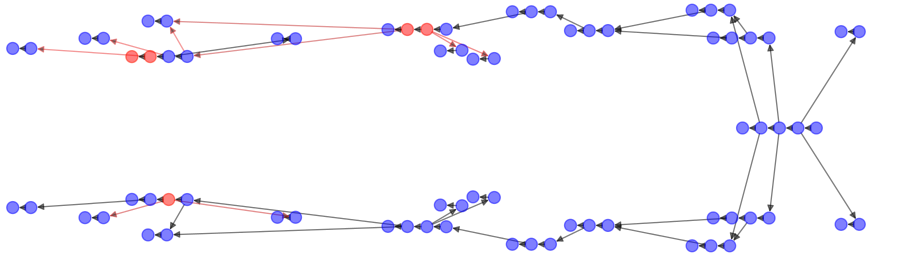}
  \end{subfigure}%
\hfill
\begin{subfigure}[b]{0.5\linewidth}
    \centering
    \includegraphics[width=\linewidth]{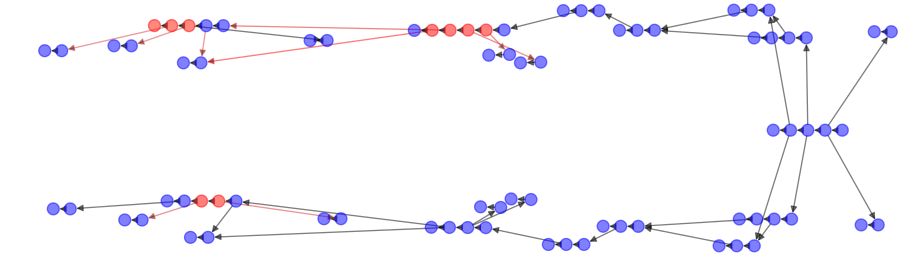}
\end{subfigure}
    \caption{mPBG segmentation differences for simulations (0) and (27) due to different times of absorption.}
    \label{fig:sgmnt}
\end{figure}

We use an illustrative example presented in \cite{pakiman2023smrnk} to evaluate our method.
This study contains 66 simulations; each model contains 27 parts and 11 components.
The model structure is the same, therefore the graph structure remains the same for all simulations.
Figure \ref{fig:extrct_g} shows the extracted graph 
for CBG, sPBG and mPBG.
Here, in the graph visualisation, the vertices are positioned in the centre of its part or component box.
In Figure \ref{fig:cmpnt} for CBG, the vertices of the graph are labelled by these components.
For sPBG and mPBG each vertex refers to a part in figures \ref{fig:open} and \ref{fig:looped},
where the parts corresponding to the vertex of a component are coloured grey. 
The mPBG has additional edges compared to sPBG that are marked in red, Figure \ref{fig:looped}.
While the CBG, sPBG and mPBG graphs are the same for 66 simulations, adding the time segmentation to the graphs can change the structure for each simulation due to different time sequences.
Figure \ref{fig:sgmnt} shows the differences in two simulations generated by time segmentation for \ac{mPBG}.
In the following sections, we evaluate the computation of the $IE$ flow and the detection of the load-path.

\vspace{-10pt}

\footnotetext{The zoomed views use \codebox{networkx.kamada_kawai_layout()} with vertex distances and positions to improve the visualisation.}

\subsection{Graph Flow}
\label{subsec:grph_flw}
We use the RSME of the inflow and outflow to evaluate the flow calculation as:
\begin{equation}\label{eq:flow1}
  RSME = \sqrt{ \frac{1}{N}\sum_{j=1}^{N} IE_j - \left(  \sum_{n \in I(j)}{w_{n, j}} - \sum_{n \in O(j)}{w_{j, n}}\right) }.
\end{equation}
The flow calculation has a small error in the order of $2$ to $3e-16$ for the three graph extraction methods.
The comparatively high spread of the RMSE for CBG indicates that for some simulations the connectivity of the CBG graph is limited, which increases the RMSE for these simulations.

\subsection{Load-path detection}
\label{subsec:ldpth}

Here, we first discuss the result of the load-path detection for five reference models, as in \cite{pakiman2023smrnk}, and show how the load-path detection characterises the simulations.
Then, we use the best method to classify all 66 simulations.
In the reference simulations -- 3, 30, 31, 60, 61 --- the crash-box thicknesses differ as follows.
Simulation 3 has the same thickness on both \ac{LHS} and \ac{RHS}.
Compared to 3, simulations 30 and 31 are less stiff on \ac{RHS} and \ac{LHS}, respectively.
Whereas simulations 60 and 61 are stiffer on \ac{LHS} and \ac{RHS}, respectively, compared to 3.

\begin{figure}[t]
  \centering
  \scriptsize

\begin{subfigure}[b]{0.06\linewidth}
  \centering
    \raisebox{-3 em}{}
\end{subfigure}%
\hfill
\begin{subfigure}[b]{0.078\linewidth}
  \centering
    \raisebox{-3 em}{CBG}
\end{subfigure}%
\hfill
\begin{subfigure}[b]{0.078\linewidth}
  \raisebox{-3 em}{sPBG}
\end{subfigure}%
\hfill
\begin{subfigure}[b]{0.078\linewidth}
  \raisebox{-3 em}{mPBG}
\end{subfigure}%
\hfill
\begin{subfigure}[b]{0.078\linewidth}
  \centering
  \raisebox{-3 em}{CBG}
\end{subfigure}%
\hfill
\begin{subfigure}[b]{0.078\linewidth}
\raisebox{-3 em}{sPBG}
\end{subfigure}%
\hfill
\begin{subfigure}[b]{0.078\linewidth}
\raisebox{-3 em}{mPBG}
\end{subfigure}%
\hfill
\begin{subfigure}[b]{0.078\linewidth}
  \centering
  \raisebox{-3 em}{CBG}
\end{subfigure}%
\hfill
\begin{subfigure}[b]{0.078\linewidth}
\raisebox{-3 em}{sPBG}
\end{subfigure}%
\hfill
\begin{subfigure}[b]{0.078\linewidth}
\raisebox{-3 em}{mPBG}
\end{subfigure}%
\hfill
\begin{subfigure}[b]{0.078\linewidth}
  \centering
  \raisebox{-3 em}{CBG}
\end{subfigure}%
\hfill
\begin{subfigure}[b]{0.078\linewidth}
\raisebox{-3 em}{sPBG}
\end{subfigure}%
\hfill
\begin{subfigure}[b]{0.078\linewidth}
\raisebox{-3 em}{mPBG}
\end{subfigure}%
\hfill

\begin{subfigure}[b]{0.064\linewidth}
        \raisebox{3 em}{(3)}
\end{subfigure}%
\hfill
\begin{subfigure}[b]{0.22\linewidth}
      \centering
        \includegraphics[width=\linewidth]{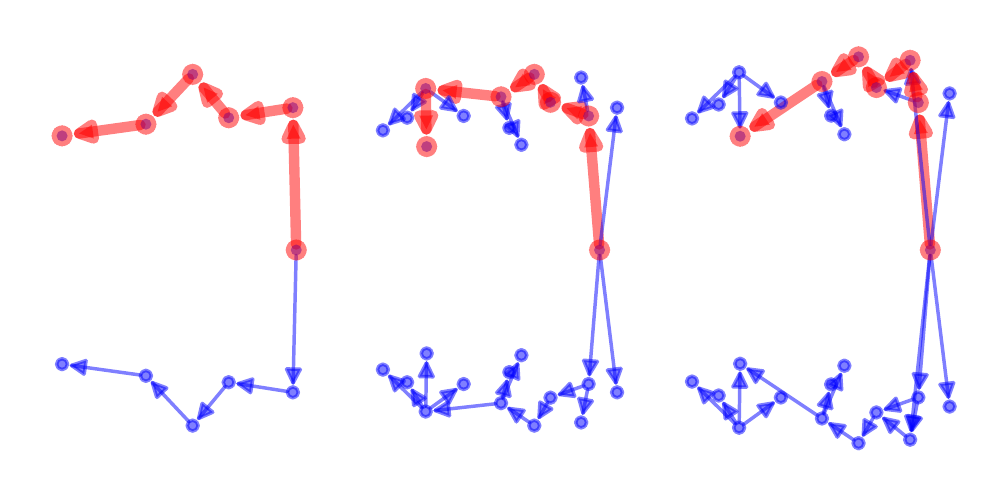
        }
        \label{fig:IE_ldpth_3}
\end{subfigure}%
\hfill
\begin{subfigure}[b]{0.22\linewidth}
      \centering
        \includegraphics[width=\linewidth]{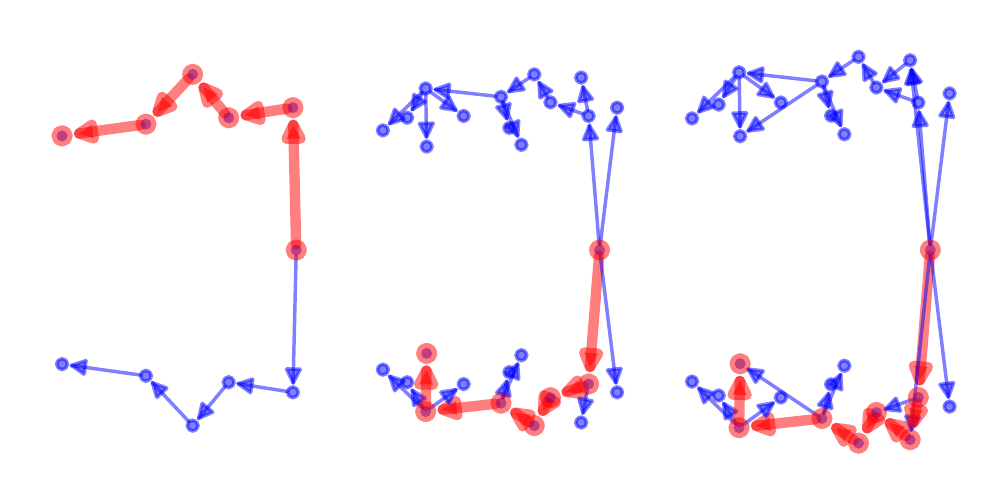
        }
        \label{fig:IE_dt_ldpth_3}
\end{subfigure}%
\hfill
  \begin{subfigure}[b]{0.22\linewidth}
      \centering
        \includegraphics[width=\linewidth]{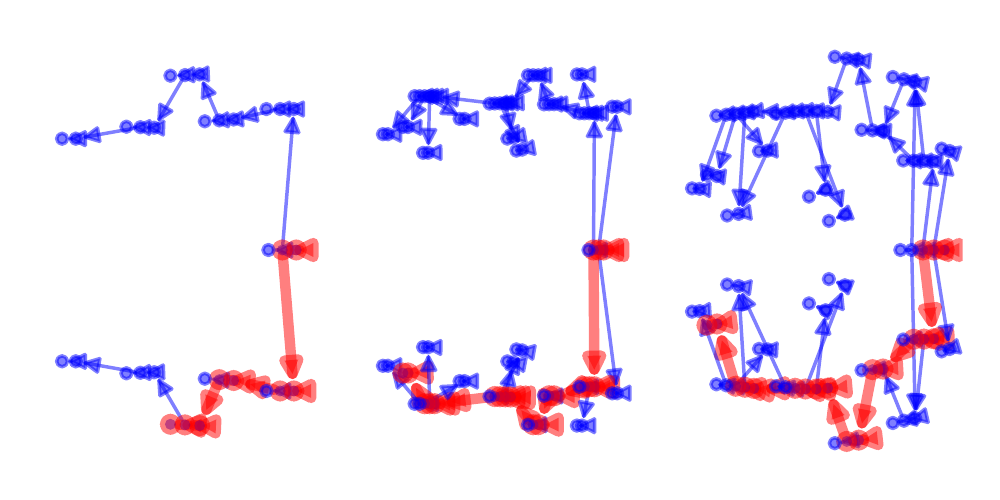
        }
        \label{fig:sgmnt_ldpth_3}
\end{subfigure}%
\hfill
\begin{subfigure}[b]{0.22\linewidth}
      \centering
        \includegraphics[width=\linewidth]{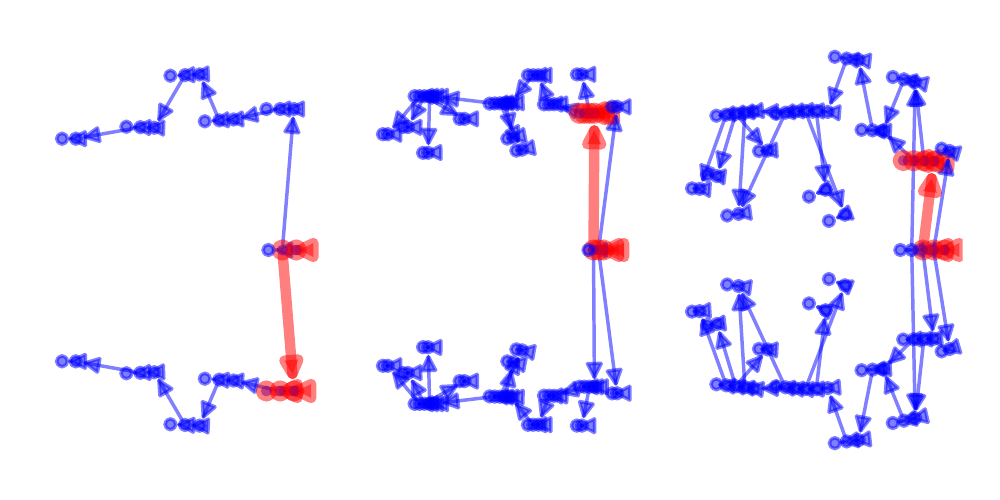}
        \label{fig:sgmnt_IEdt_ldpth_3}
\end{subfigure}%
\hfill
\begin{subfigure}[b]{0.064\linewidth}
      \raisebox{3 em}{(30)}
\end{subfigure}%
\hfill
\begin{subfigure}[b]{0.22\linewidth}
      \centering
        \includegraphics[width=\linewidth]{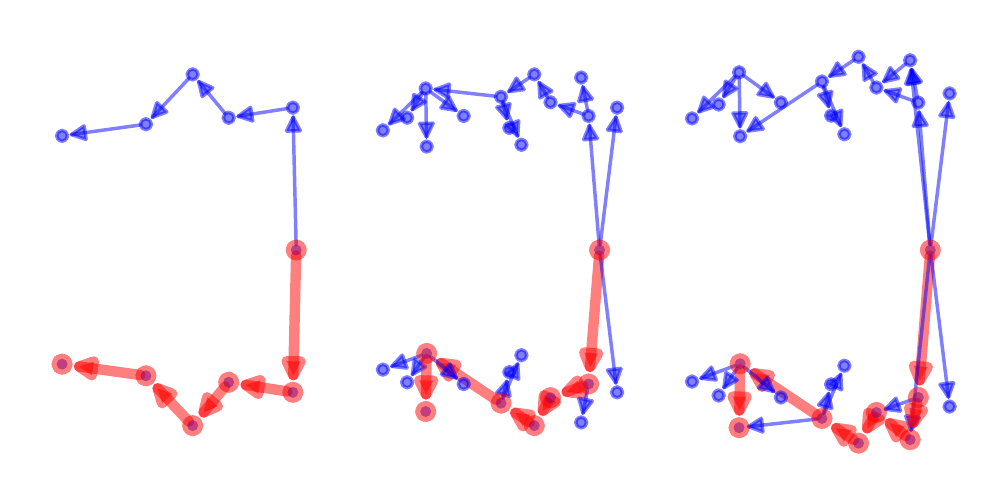
        }
        \label{fig:IE_ldpth_30}
\end{subfigure}%
\hfill
\begin{subfigure}[b]{0.22\linewidth}
      \centering
        \includegraphics[width=\linewidth]{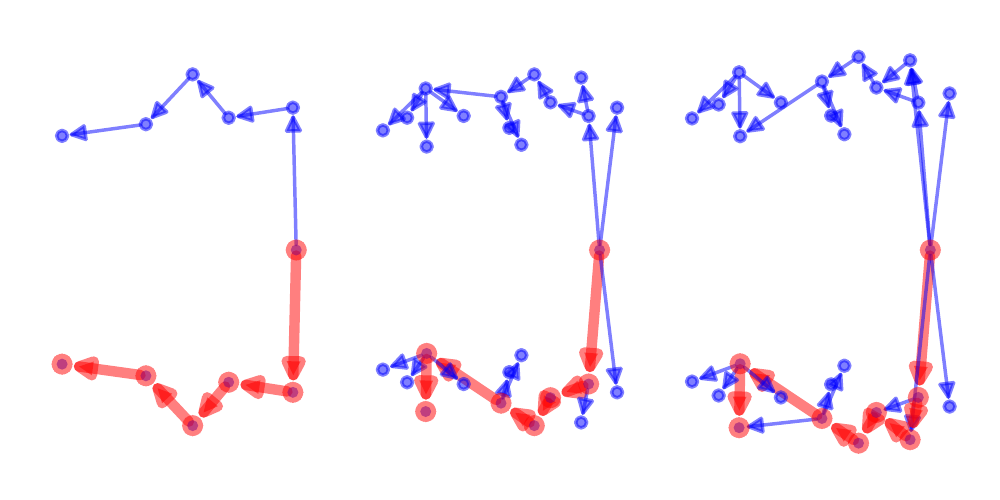
        }
        \label{fig:IE_dt_ldpth_30}
\end{subfigure}%
\hfill
\begin{subfigure}[b]{0.22\linewidth}
    \centering
    \includegraphics[width=\linewidth]{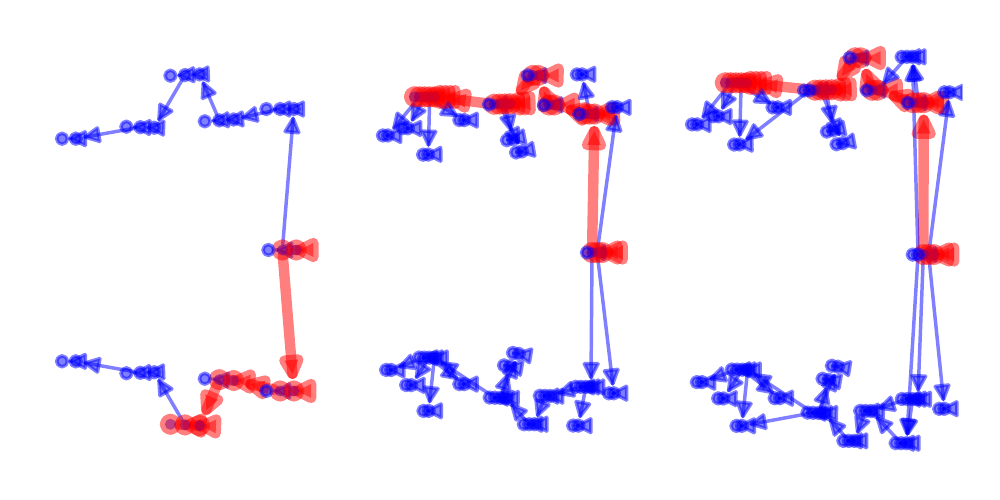}
      \label{fig:sgmnt_ldpth_30}
\end{subfigure}%
\hfill
\begin{subfigure}[b]{0.22\linewidth}
      \centering
        \includegraphics[width=\linewidth]{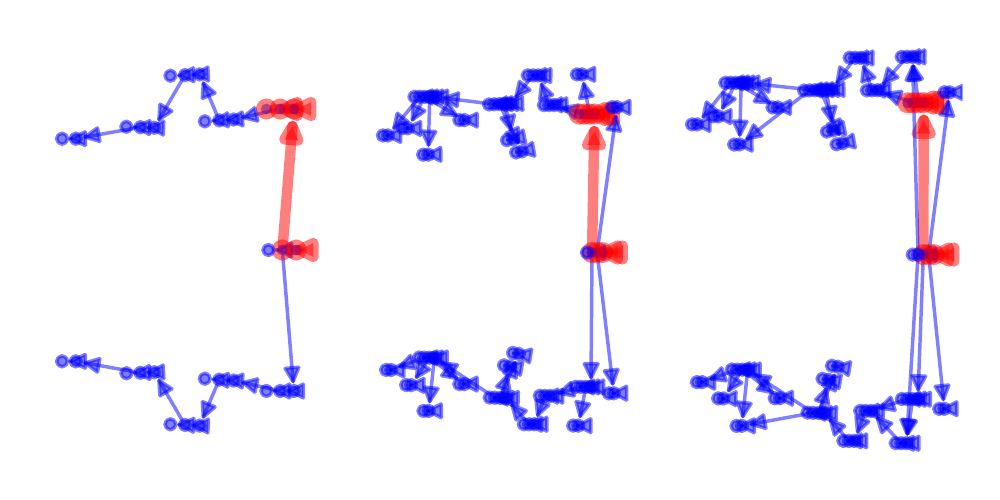}
        \label{fig:sgmnt_IEdt_ldpth_30}
\end{subfigure}%
\hfill
\begin{subfigure}[b]{0.064\linewidth}
      \raisebox{3 em}{(31)}
\end{subfigure}%
\hfill
\begin{subfigure}[b]{0.22\linewidth}
  \centering
    \includegraphics[width=\linewidth]{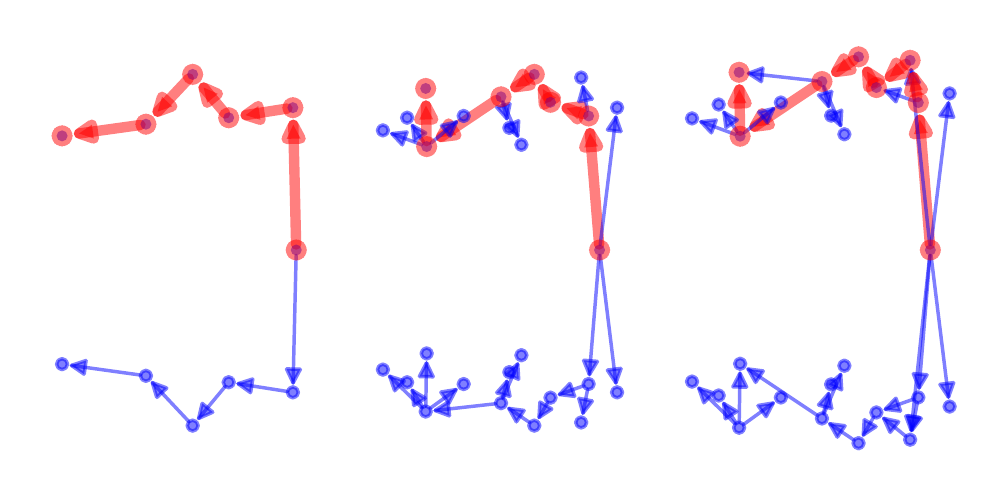
    }
    \label{fig:IE_ldpth_31}
\end{subfigure}%
\hfill
\begin{subfigure}[b]{0.22\linewidth}
      \centering
        \includegraphics[width=\linewidth]{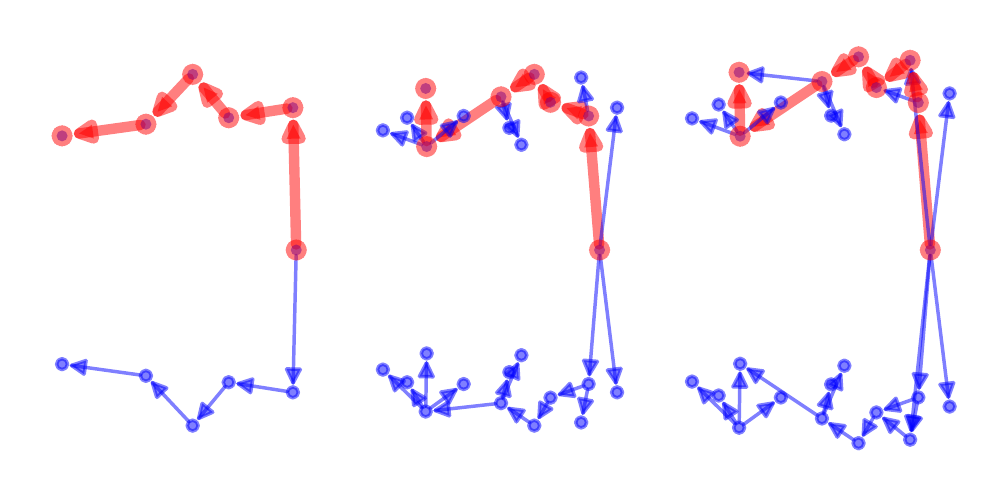
        }
        \label{fig:IE_dt_ldpth_31}
\end{subfigure}%
\hfill
\begin{subfigure}[b]{0.22\linewidth}
  \centering
  \includegraphics[width=\linewidth]{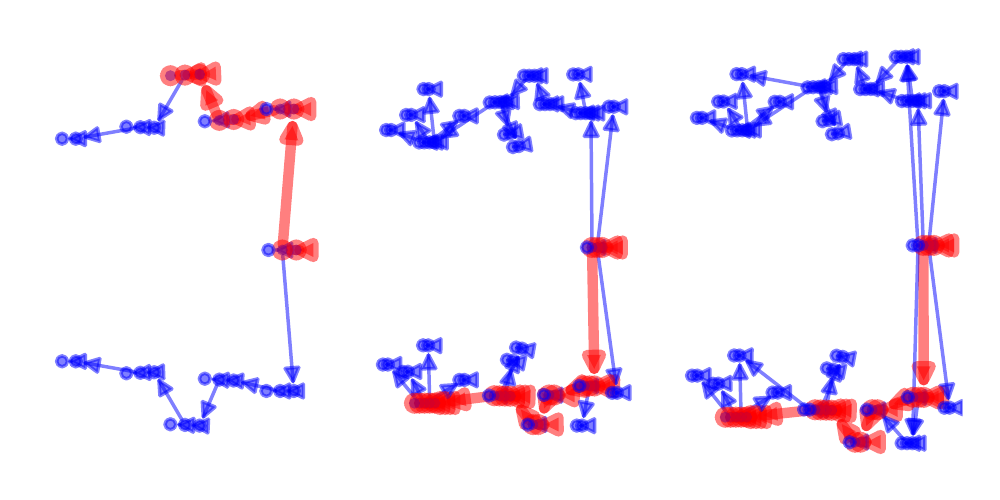}
    \label{fig:sgmnt_ldpth_31}
  \end{subfigure}%
\hfill
\begin{subfigure}[b]{0.22\linewidth}
      \centering
        \includegraphics[width=\linewidth]{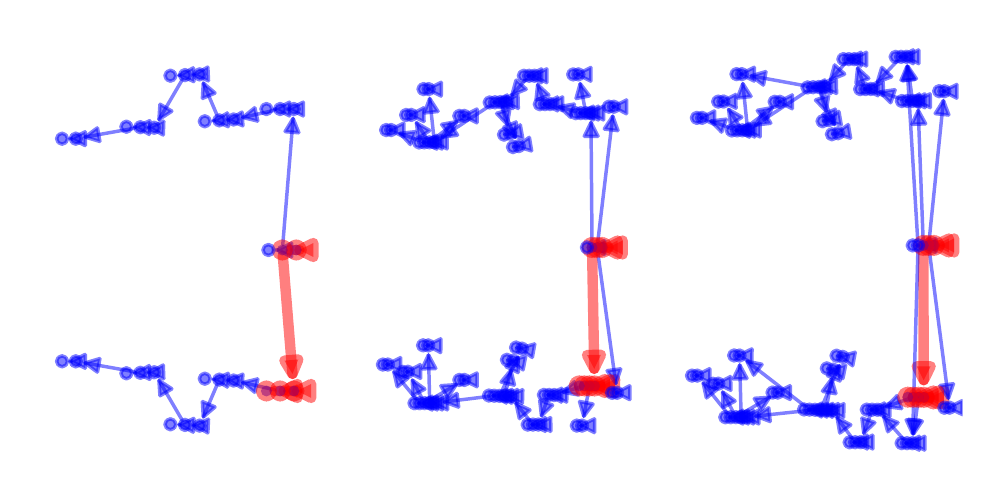}
        \label{fig:sgmnt_IEdt_ldpth_31}
\end{subfigure}%
\hfill
\begin{subfigure}[b]{0.064\linewidth}
      \raisebox{3 em}{(60)}
\end{subfigure}%
\hfill
\begin{subfigure}[b]{0.22\linewidth}
    \centering
      \includegraphics[width=\linewidth]{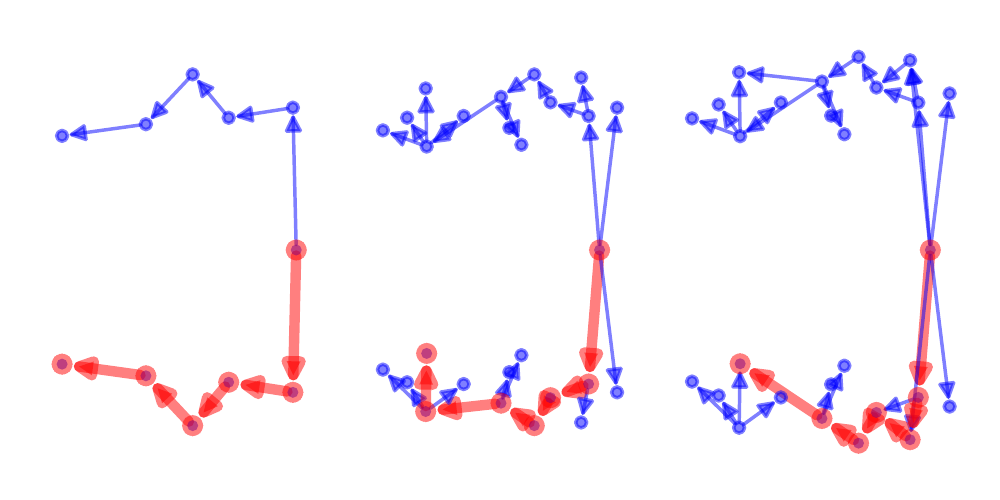
      }
      \label{fig:IE_ldpth_60}
\end{subfigure}%
\hfill
\begin{subfigure}[b]{0.22\linewidth}
      \centering
        \includegraphics[width=\linewidth]{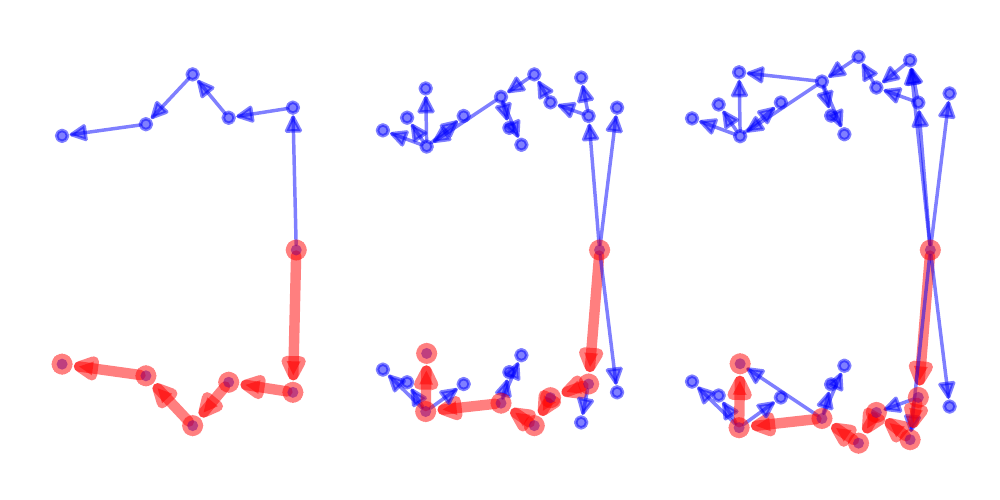
        }
        \label{fig:IE_dt_ldpth_60}
\end{subfigure}%
\hfill
  \begin{subfigure}[b]{0.22\linewidth}
      \centering
      \includegraphics[width=\linewidth]{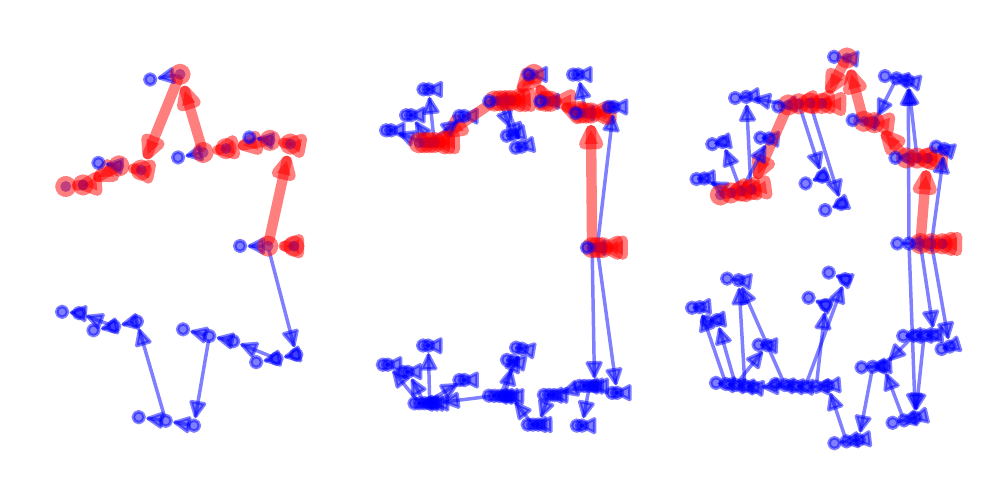}
        \label{fig:sgmnt_ldpth_60}
  \end{subfigure}%
\hfill
\begin{subfigure}[b]{0.22\linewidth}
      \centering
        \includegraphics[width=\linewidth]{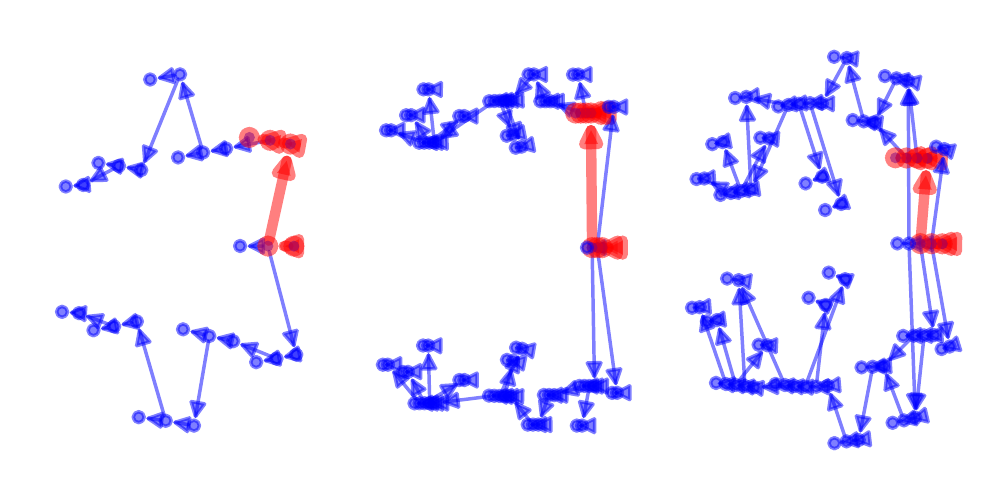}
        \label{fig:sgmnt_IEdt_ldpth_60}
\end{subfigure}%
\hfill
\begin{subfigure}[b]{0.06\linewidth}
      \raisebox{5 em}{(61)}
\end{subfigure}%
\begin{subfigure}[b]{0.22\linewidth}
      \centering
        \includegraphics[width=\linewidth]{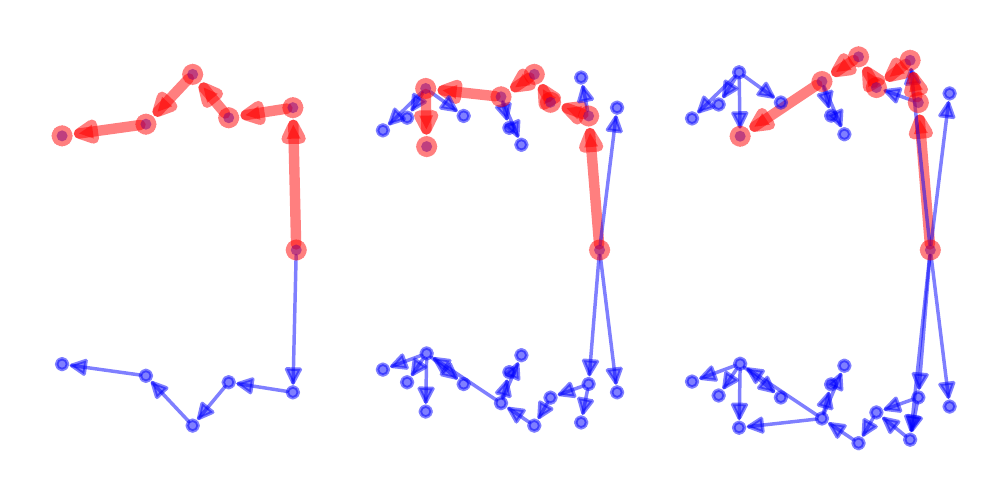
        }
        \caption{$f_{IE}$}
        \label{fig:IE_ldpth_61}
  \end{subfigure}%
\hfill
  \begin{subfigure}[b]{0.22\linewidth}
        \centering
          \includegraphics[width=\linewidth]{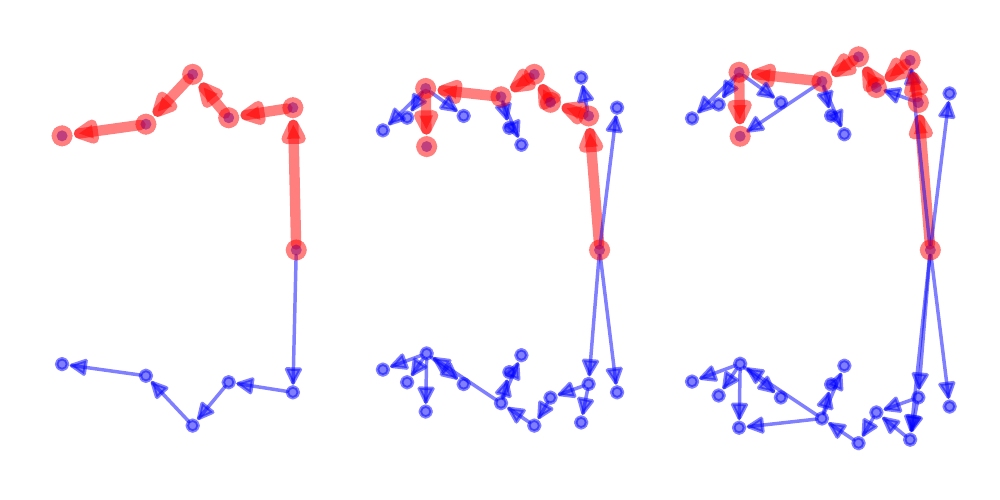
          }
          \caption{$f_{IE\Delta t}$}
          \label{fig:IE_dt_ldpth_61}
  \end{subfigure}%
  \hfill
      \begin{subfigure}[b]{0.22\linewidth}
          \centering
          \includegraphics[width=\linewidth]{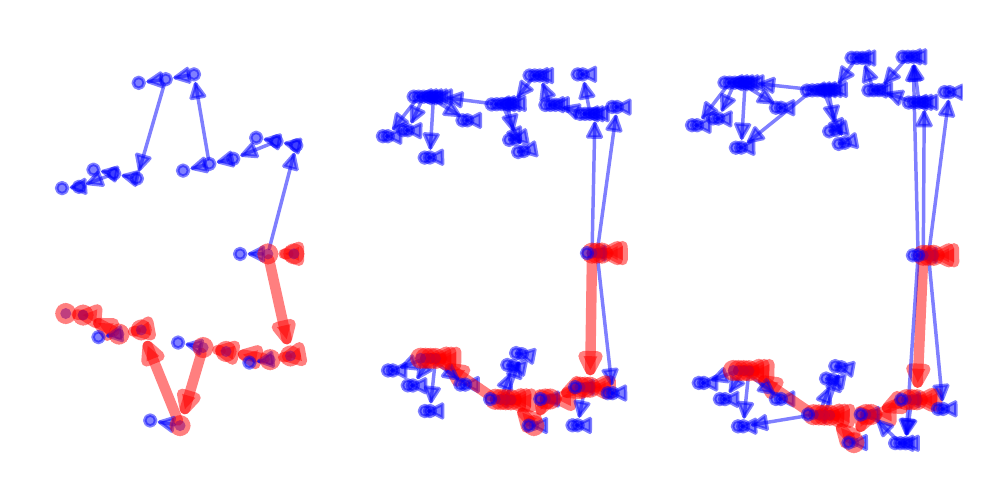}
            \caption{$s_t$}
            \label{fig:sgmnt_ldpth_61}%
    \end{subfigure}%
  \hfill
  \hfill
  \begin{subfigure}[b]{0.22\linewidth}
        \centering
          \includegraphics[width=\linewidth]{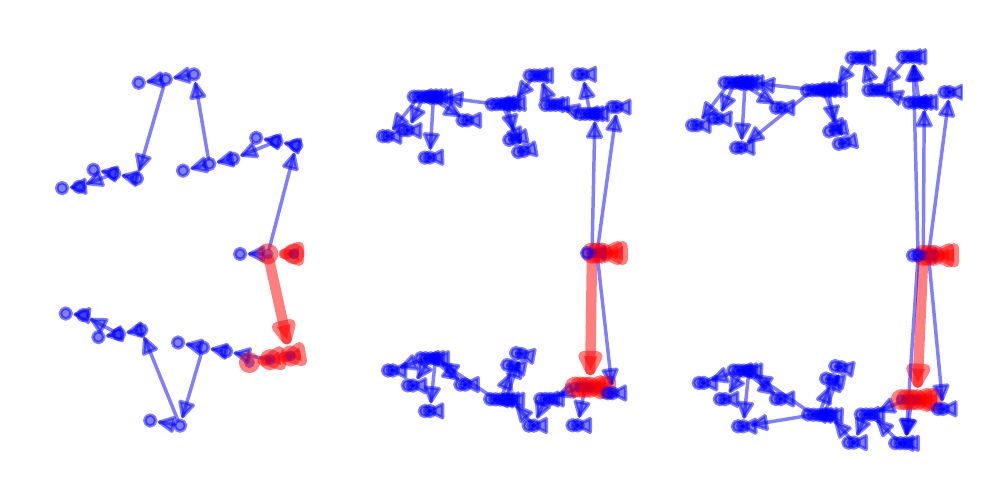}
          \caption{$s_{P_e}$}
          \label{fig:sgmnt_IEdt_ldpth_61}
  \end{subfigure}%
    \caption{
    Load-path detection, marked in red, for the five reference simulations from~\cite{pakiman2023smrnk}. \ac{LHS} and \ac{RHS} are at the top and bottom, respectively.
    Each simulation and edge weight setup include results of CBG, sPBG and mPBG.
    }
    \label{fig:ldpth}
\end{figure}

Figure \ref{fig:ldpth} summarises the load-path detection with four edge weights as described in Section \ref{sec:lp_dtc}.
Columns \subref{fig:IE_ldpth_61} and \subref{fig:sgmnt_ldpth_61} are the single feature results for $f_{IE}$ and $s_t$.
The other two columns are weighted with combined features $f_{IE\Delta t}$ and $s_{P_e}$, columns \subref{fig:IE_dt_ldpth_61} and \subref{fig:sgmnt_IEdt_ldpth_61} respectively.
We show the results of three different graph extraction methods for each scenario and the detected paths are marked in red.
Based on the structural stiffness, the expected energy load-path for simulations 30 and 60 is at the \ac{RHS} (bottom) and for simulations 31 and 61 at the \ac{LHS} (top).

We expect that for graphs with $f_{IE}$-edge weight, it is the reverse of graphs with $s_t$-weight whether we get a top or bottom load-path.
This is due to the physics of the problem, i.e. stiffer parts take more time for absorption and deform less, which means lower $IE$.
The only exception we observe is in the result with CBG and $s_t$ weighting.
Here the detected path for these simulations does not continue to the side-member and a different side of the structure is detected compared to sPBG or mPBG.
This example shows the limitation of CBG in time feature extraction, i.e., the component level is less sensitive than the part level.

Next for the combined features, $f_{IE\Delta t}$ and $s_{P_e}$, for most scenarios the detected load-path remains in the expected direction of the structure.
The only exception is the CBG graph for simulation 3.
Simulation 3 is a symmetric model and lacks a dominant load-path due to its symmetry.
Again, the CBG method lacks the detail to realise the effect of time in detecting the load-path.
The additional obvious observation is that with $s_{P_e}$ weight the detected path is shorter.
This detection describes well that the crash-box influence is much greater than those of the remaining parts.
Therefore, this path captures the efficient path of the load rather than the full path along the structure.

\begin{figure}[t]
  \centering
    \includegraphics[width=.85\linewidth]{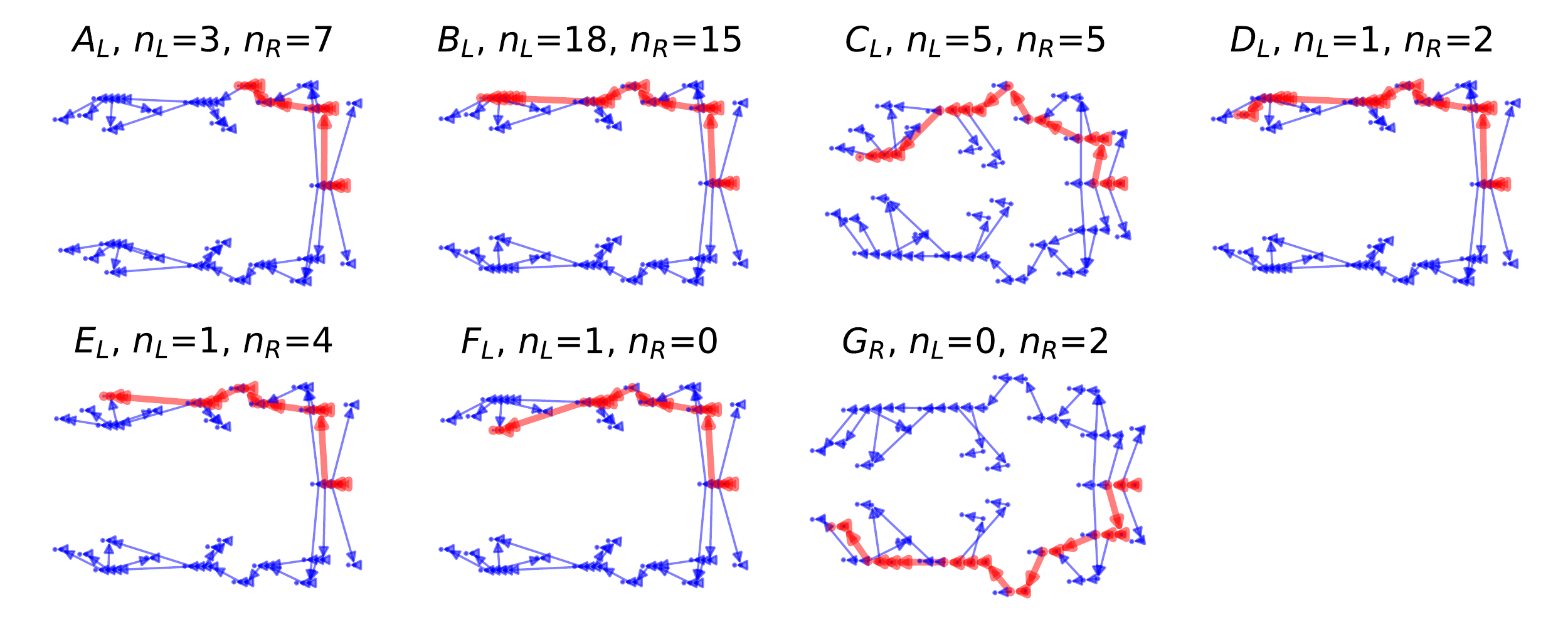}
    \caption{Identical load-paths, marked in red, that are identified for the simulation dataset from~\cite{pakiman2023smrnk}. Only one is shown if there is a symmetric pair. $n_L$ and $n_R$ are the number of occurrences of a load-path in the dataset, respectively.
    }
    \label{fig:sgmnt_t_all}
\end{figure}

Among these approaches, the mpBG with $s_t$ detects the most detailed load-paths, which is better for simulation comparison.
As a result, we use it to visually categorise all 66 simulations.
This method categorises the data into 12 identical load-paths, where 10 are symmetric pairs, i.e., an LHS path corresponds to an RHS path.
Figure \ref{fig:sgmnt_t_all} summarises the clusters.
Most of the simulations, 33, are grouped in cluster B.
The biggest difference of the clusters is between cluster A and the rest where the path ends with a crash-box absorption.
The remaining clusters have similar absorption for the crash-box and differ in vertex selection for the side-member at the end of the path.

%
%
\section{Conclusion and Outlook}
\label{sec:conclusion}

We considered load-path detection in crash analysis, one of the automotive \ac{CAE} domains, by using graph approaches.
Due to the lack of graphs in the \ac{CAE} data, we introduced graph extraction methods to convert the \ac{CAE} analysis of crashes into graphs.
To characterise the absorption path of the vehicle structure, we not only abstract the vehicle structure into the graph, but also define edge directions and edge weights.
By computing the longest weighted path in a graph an automated detection of load-paths now becomes feasible.
Vehicles with the same structural design have an almost similar graph structure, while edge weighting and time segmentation detect differences in load-paths.
Our method showed promising results analysing an illustrative example with 66 simulations.
Based on our study, it is best to use different graph extraction approaches and edge weights ($w$) for different applications, as follows:
\begin{enumerate}[label=\alph*]
  \item CBG, $w = f_{IE}$: crash mode analysis \cite{pakiman2023smrnk}, advantage: simple and stable.
  \item mPBG, $w = f_{IE\Delta t}$: $IE$ flow path analysis, advantage: more details.
  \item mPBG, $w = s_t$: simulation clustering using load-path, advantage: sensitivity to time sequence.
  \item mPBG $w = s_{P_e}$: analyse part or component efficiency.
\end{enumerate}

As well as being useful for the \ac{CAE} engineers, the load-path clusters from c) can also be used as labels, which opens up new possibilities for using supervised \ac{ML} for \ac{CAE}.
We see as a next stage an implementation of graph embedding methods to automatically classify the results.

In addition, posture detection methods can be used to further process the data during the crash \cite{ma2022survey}.
With these methods, part features should remain at the vertex level for active part detection.
However, as far as we are aware, there is limited research on directed graphs to find the load-path.
Furthermore, converting a whole vehicle into a graph requires additional considerations.
For a complete vehicle, graph extraction can often lead to several unconnected graphs due to the existence of larger parts.
Our graph extraction works for sub-models, but further heuristics are needed to extend its application, which is beyond the scope of this work.
Finally, we extracted a static graph from the undeformed geometry.
As the deformed structure may lead to additional contacts between parts that do not exist in the undeformed structure, it may be useful in the future to consider the deformed structures as well.

\begin{acronym}
    \acro{FEM}{finite element method}
\acro{FE}{finite element}
\acro{CAD}{computer aided design}
\acro{CAE}{Computer aided engineering}
\acro{GML}{graph machine learning}
\acro{ML}{machine learning}
\acro{$IE$}{internal energy}
\acro{$IE_c$}{component internal energy}
\acro{$KE$}{kinetic energy}
\acro{$t_n$}{final absorption time}
\acro{$t_i$}{initial absorption time}
\acro{$IE_{max}$}{internal energy maximum}
\acro{ATE}{absorption time efficiency}

\acro{LHS}{left hand side}
\acro{RHS}{right hand side}
\acro{c-RHS}{crash-box RHS}
\acro{c-LHS}{crash-box LHS}
\acro{b}{bumper-beam}
\acro{s-LHS}{side-member LHS}
\acro{s-RHS}{side-member RHS}
\acro{COG}{center of gravity}

\acro{KG}{knowledge graph}
\acro{DSM}{deformation space models}
\acro{LMS}{lumped mass spring}

\acro{CBG}{component-based graph}
\acro{sPBG}{single part-based graph}
\acro{mPBG}{multi part-based graph}

\end{acronym}

\bibliographystyle{ieeetr}
\bibliography{00_physical_graph}

\end{document}